\documentclass[letterpaper, 10pt, conference]{ieeeconf}
\IEEEoverridecommandlockouts                             
\usepackage{graphics} % for pdf, bitmapped graphics files
\usepackage{epsfig} % for postscript graphics files
\usepackage{times} % assumes new font selection scheme installed
\usepackage{amsmath,mathtools} % assumes amsmath package installed
\usepackage{yhmath}
\usepackage{amssymb}  % assumes amsmath package installed
\usepackage{centernot}
\usepackage{pbox}
\usepackage{svg}
\usepackage{bm}
\usepackage[most]{tcolorbox}
\usepackage{xcolor} % for color definitions

\usepackage{amsfonts}

\usepackage{xcolor}
\usepackage{listings}
\usepackage{url}
\usepackage{hyperref}

\usepackage{xcolor}

\usepackage{tikz}       % for \foreach
\usepackage[]{graphicx} % use draft option to reduce compile times
\usepackage[export]{adjustbox}
\usepackage{svg}
\usepackage{subcaption}
\usepackage{caption}
\usepackage{dsfont}

\usepackage{color, colortbl}
\definecolor{Gray}{gray}{0.9}
\usepackage{xspace}
\usepackage{tcolorbox}

\usepackage{cite}
\usepackage{amsfonts}
\usepackage{bm}
\usepackage{breqn}

\usepackage{diagbox}
\usepackage{multirow}
\usepackage{multicol}
\usepackage{graphicx}
 \usepackage[subtle,tracking=normal]{savetrees}

\usepackage{balance}
\usepackage{caption}
\usepackage{subcaption}
\usepackage{makecell}
\usepackage{microtype}
\usepackage{indentfirst}
\usepackage{pdfpages}
\usepackage[mode=match]{siunitx}

\usepackage{colortbl}
\usepackage{xcolor}
\usepackage{makecell} % for \thead

\definecolor{cellred}{RGB}{255,150,150}
\definecolor{cellorange}{RGB}{255,210,150}
\definecolor{cellgreen}{RGB}{170,255,170}

\usepackage[dvipsnames]{xcolor}
\definecolor{ao(english)}{rgb}{0.0, 0.5, 0.0}
\definecolor{green(pigment)}{rgb}{0., 0.75, 0.2}
\usepackage{soul}

\usepackage{microtype}
\usepackage{indentfirst}
\usepackage{pdfpages}
\usepackage[mode=match]{siunitx}
\usepackage{hyperref}
\usepackage[percent]{overpic}
\usepackage{xcolor}

\newcounter{messagecounter}
\newcommand{\messagecounter}{\arabic{messagecounter}}

\newif\ifanonymous
\title{\LARGE \bf
PGTT: Phase-Guided Terrain Traversal for Perceptive Legged Locomotion
}

\ifanonymous
\author{Anonymous Authors$^1$%
\thanks{$^{1}$ Anonymous Affiliation}}
\else
\author{Alexandros Ntagkas$^{1,2}$, Chairi Kiourt$^{2,3}$, and Konstantinos Chatzilygeroudis$^{1,2,4}$% <-this % stops a space
\thanks{*This work was supported by the Hellenic Foundation for Research and Innovation (H.F.R.I.) under the ``3rd Call for H.F.R.I. Research Projects to support Post-Doctoral Researchers'' (Project Acronym:~NOSALRO, Project Number:~7541). This work has also been partially supported by project MIS 5154714 of the National Recovery and Resilience Plan Greece 2.0 funded by the European Union under the NextGenerationEU Program.}% <-this % stops a space
\thanks{$^{1}$Laboratory of Automation and Robotics (LAR) in the Department of Electrical \& Computer Engineering,
        University of Patras, GR-26504 Patras, Greece,
        {\tt\small a\_ntagkas@ac.upatras.gr, costashatz@upatras.gr}}%
\thanks{$^{2}$Archimedes/Athena RC, Greece}%
\thanks{$^{3}$Athena - Research and Innovation Center in Information, Communication and Knowledge Technologies, Xanthi, Greece,
        {\tt\small chairiq@athenarc.gr}}
\thanks{$^{4}$Computational Intelligence Laboratory (CILab), Department of Mathematics, University of Patras, GR-26110 Patras, Greece}%
\vspace{-3em}
}
\fi

\ifanonymous
\newcommand{\CodeURL}{https://anonymous.4open.science/status/phase_guided_terrain_traversal-1B1D}
\else
\newcommand{\CodeURL}{https://github.com/NtagkasAlex/phase_guided_terrain_traversal}
\fi

\begin{document}
\maketitle
\thispagestyle{empty}
\pagestyle{empty}

\begin{abstract}

State-of-the-art perceptive Reinforcement Learning controllers for legged robots typically either (i) impose oscillator- or IK-based gait priors that constrain the action space, bias policy optimization, and limit adaptability across robot morphologies, or (ii) operate ``blind,'' making them unable to anticipate hind-leg terrain and brittle to observation noise.
We propose \textbf{Phase-Guided Terrain Traversal (PGTT)}, a perception-aware deep-RL \emph{approach} that enforces gait structure through reward shaping, thereby \emph{reducing inductive bias} compared to oscillator- or IK-conditioned action priors.
PGTT encodes per-leg phase as a cubic Hermite spline, adapts swing height to local heightmap statistics, and adds a swing-phase contact penalty, while the policy acts directly in joint space for morphology-agnostic deployment.
Trained in MuJoCo (MJX) on procedurally generated stair-like terrains with curriculum learning and domain randomization, PGTT achieves the highest success rate among the evaluated baselines under push disturbances (median +7.5\% over the next-best baseline) and on discrete obstacles (+9\%), while maintaining comparable velocity tracking.
We validate PGTT on a Unitree Go2 using a real-time LiDAR elevation-to-heightmap pipeline and report preliminary results on \textbf{ANYmal-C} using the same hyperparameters.
These results provide early evidence that terrain-adaptive, phase-guided reward shaping can transfer across platforms without platform-specific policy priors or extensive retuning. 

\end{abstract}

\section{Introduction}\label{sec:introduction}
Legged robots promise unmatched mobility in cluttered, uneven, and human‑made environments, but robust gait control on such terrain remains challenging~\cite{Hutter2017Anymal,Kuindersma2016Atlas}. Reinforcement learning (RL) has shown that agile locomotion behaviors can be learned from data~\cite{Kumar2021RMA}, yet many studies assume \emph{idealized sensing} (privileged terrain information) or operate ``blind,'' which hinders anticipation of obstacles and reduces reliability on hardware~\cite{Lee2020Learning,Duan2021Biped}. As a result, perception is essential, but the representation and how it interfaces with control are pivotal for generality and robustness.

Recent efforts incorporate visual or range sensing into the loop. Egocentric depth cameras enable end‑to‑end training and have demonstrated stair and gap traversal~\cite{Agarwal2022Legged}, but limited field of view, sensor noise, and the need for temporal memory remain practical obstacles (especially for hind‑leg terrain). Methods that rely on globally consistent elevation maps or multi‑sensor rigs can extend foresight but require careful calibration and accurate global pose estimation, which can be brittle~\cite{Miki2022Expanding}. In parallel, many RL controllers encode \emph{gait priors} by prescribing oscillatory foot/joint targets as functions of a per‑leg phase and tracking them with inverse kinematics (IK) and PD controllers; while effective, this constrains the action space and couples policies to specific morphologies, introducing bias that may reduce adaptability~\cite{Lee2020Learning,Miki2022}.

We propose \textbf{Phase‑Guided Terrain Traversal (PGTT)}, a perception‑aware deep‑RL approach that retains the benefits of rhythmic structure while avoiding IK and action‑space constraints. PGTT uses a robot‑centric \emph{heightmap} (derived online from LiDAR elevation mapping) as a compact terrain representation and encodes per‑leg phase with a \emph{cubic Hermite spline} whose swing apex adapts to local height statistics. Crucially, the phase prior is enforced \emph{only through reward shaping}, while the policy acts directly in joint space. This design keeps the action space unconstrained and \emph{reduces inductive bias} in policy learning compared to oscillator/IK‑conditioned targets, easing deployment across different morphologies.%; see the pipeline in Fig.~4 and the reward terms in Table~I.

\begin{figure}[t]
    \centering
    \includegraphics[width=0.9\linewidth]{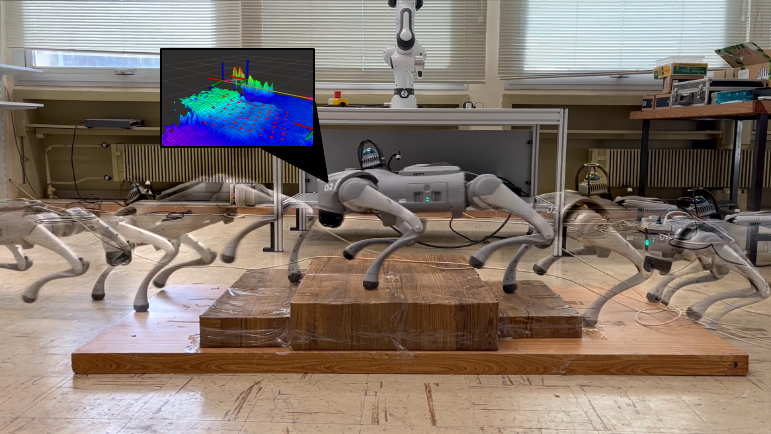}
    \caption{Real‑world example of the Unitree Go2 robot climbing stair terrain.}
    \vspace{-2em}
\end{figure}

Across extensive simulation tests, PGTT exhibits faster learning convergence than an end‑to‑end, non‑prior RL baseline and delivers consistently higher survival rates under pushes and on discrete obstacles than state‑of‑the‑art perceptive locomotion methods. The gains in survival come without sacrificing tracking quality, as velocity‑tracking performance remains comparable to the strongest baselines. We then deploy the learned policy on a Unitree Go2 with a real‑time elevation‑to‑heightmap pipeline and demonstrate robust stair and obstacle traversal, and we present preliminary results on \textbf{ANYmal~C} obtained \emph{without changing the hyper‑parameters}. These findings support the central premise of phase-guided reward shaping: it provides useful structure for learning while keeping the action space unconstrained, with the ANYmal-C results offering early, qualitative support for adaptability across platforms.

The main contributions of this manuscript are:
\begin{itemize}
    \item A terrain‑adaptive, phase‑guided \emph{reward} that encodes a Hermite‑spline swing trajectory driven by local heightmap statistics and penalizes swing‑phase contacts, without constraining the action space or using IK, thereby reducing inductive bias and improving morphology‑agnostic deployment.
    \item A compact perception‑to‑policy design that feeds a robot‑centric heightmap directly to the policy, avoiding global pose assumptions while capturing nearby terrain geometry relevant for stepping~\cite{Duan2021Biped}.
    \item An extensive evaluation on stair‑like and discrete‑obstacle terrains showing higher success rates, Sim2Real transfer on a Go2 quadruped, and preliminary cross‑platform results on ANYmal~C with zero hyper‑parameter changes.
    \item An accessible training stack using MuJoCo/MJX that provides accurate dynamic simulation and high throughput on a single consumer GPU, offering a lightweight alternative to Isaac Gym‑based pipelines~\cite{makoviychuk2021isaacgymhighperformance}\footnote{Our code is available at \url{\CodeURL}.}.
\end{itemize}

\begin{figure*}[t]
    \centering
    \includegraphics[width=\textwidth]{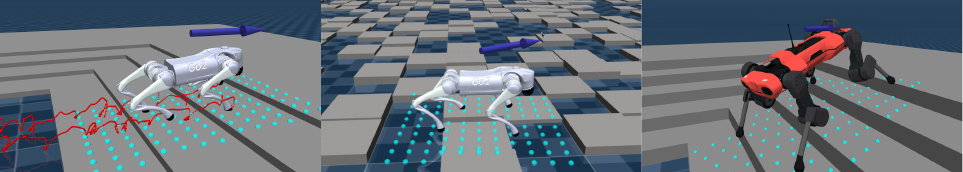}
    \caption{Simulation snapshots of PGTT. Left: Go2 on stairs with projected front‑foot trajectories (red). Middle: Go2 traversing discrete obstacles. Right: ANYmal~C on stairs.}
    \label{fig:simulation}
    \vspace{-2em}
\end{figure*}

\section{Related Work}\label{sec:related}

Learning‑based locomotion has evolved along two main lines. In one, a learned module proposes high‑level commands such as footholds or body motion that a model‑based controller then tracks; examples include RLOC and hierarchical vision‑in‑the‑loop architectures that marry learned planners with Whole Body Controllers or Model Predictive Control for tracking~\cite{Gangapurwala2022,Yu2022}. In the other, policies act end‑to‑end on the robot—commanding torques or joint targets directly, and can discover strategies that hand‑crafted pipelines might miss~\cite{Lee2020Learning,Miki2022}. While hierarchical schemes benefit from explicit feasibility handling in the low‑level controller, they rely on accurate modeling and foothold execution; direct policies reduce hand‑engineering but require careful reward design and curricula for stability and transfer.

Perception choices strongly affect both families. Early ``blind'' approaches (proprioception only) showed surprising robustness but lack anticipation of obstacles~\cite{Lee2020Learning,Duan2021Biped}. Egocentric depth cameras enable end‑to‑end perception and have demonstrated stair and gap traversal, yet their limited field of view and sensor noise burden the policy with temporal memory, especially for hind‑leg terrain~\cite{Agarwal2022Legged}. Other methods depend on globally consistent elevation maps or multi‑sensor rigs; these increase foresight but demand careful calibration and accurate global pose estimation, which can be brittle in practice~\cite{Miki2022Expanding}. A robot‑centric \emph{local heightmap} offers a compact, task‑aligned alternative that captures nearby terrain geometry relevant to all legs without global consistency assumptions~\cite{Duan2021Biped}; several recent systems, including ours, adopt this representation to couple perception more tightly to control.

Phase‑augmented controllers specify foot or joint targets as functions of a per‑leg phase and track them with IK/PD controllers, improving stability but coupling the policy to morphology and introducing action‑space bias~\cite{Lee2020Learning,Miki2022}. Central Pattern Generators (CPG)-based methods similarly embed oscillators and let RL modulate their parameters, inheriting the same limitations~\cite{Bellegarda2022}. An alternative is to encode gait regularity in the \emph{objective} rather than in the actions: phase‑guided reward shaping encourages desired swing/stance timing and foot clearance while leaving the policy free to decide the final commands~\cite{Shao_2022}. This shift reduces inductive bias and eases deployment across platforms with different kinematics.

The training substrate also varies. GPU‑accelerated simulators such as Isaac Gym have popularized massively parallel data collection for on‑policy RL~\cite{makoviychuk2021isaacgymhighperformance}, while the MuJoCo/MJX stack offers practical contact simulation within a lightweight, accessible toolchain, along with good throughput for perception-aware policies. These choices interact with observation design and reward shaping: compact robot‑centric inputs and structured, but not action‑constraining, objectives typically reduce training instabilities and simplify transfer.

In this landscape, PGTT aligns with direct joint‑space control but differs in how structure is injected: it uses a robot‑centric heightmap for perception and enforces a \emph{terrain‑adaptive, phase‑guided prior purely through reward shaping}, avoiding oscillators and IK. This design aims to retain the benefits of rhythmic organization while minimizing action‑space constraints, thereby reducing inductive bias and supporting morphology‑agnostic deployment relative to oscillator/IK‑conditioned policies~\cite{Lee2020Learning,Miki2022,Shao_2022}.

\section{Phase‑Guided Terrain Traversal}
\begin{figure*}[h]
    \centering
    \includegraphics[width=0.9\textwidth]{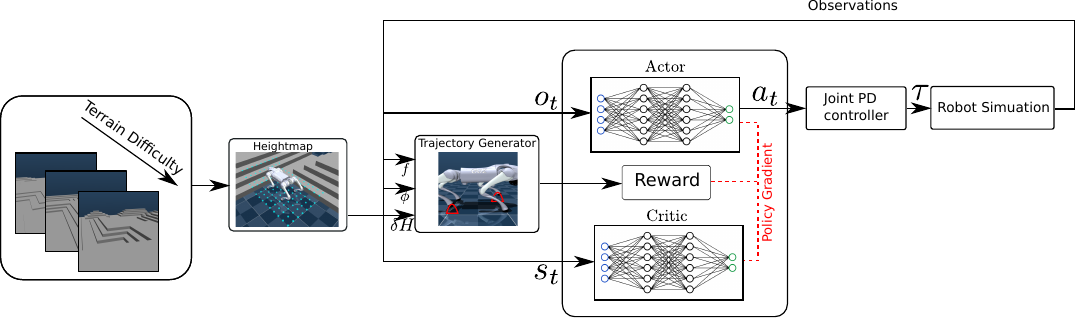}
    \caption{PGTT combines curriculum learning, a robot-centric heightmap, reward shaping through Hermite splines, asymmetric actor-critic learning, and low-level PD controllers for effective perceptive legged locomotion.}
    \label{fig:pgtt}
    \vspace{-2em}
\end{figure*}

% \subsection{Overview}
At a high level, \textbf{Phase-Guided Terrain Traversal (PGTT)} combines three ideas (Fig.~\ref{fig:pgtt}):  
(i) a compact perception module that encodes terrain as a robot-centric heightmap derived online from LiDAR measurements,  
(ii) phase variables and reward function that provide rhythmic structure without constraining the action space, and  
(iii) an asymmetric actor-critic architecture trained with PPO in GPU-accelerated MuJoCo (MJX) environments.

\subsection{Problem Formulation}
We model legged locomotion as an infinite‑horizon partially observable Markov decision process (POMDP), $\mathcal{M}=(\mathcal{S},\mathcal{A},\mathcal{O}, P, \Omega, r, \gamma, \rho_0)$,
% \[
% \mathcal{M}=(\mathcal{S},\mathcal{A},\mathcal{O}, P, \Omega, r, \gamma, \rho_0),
% \]
where $s_t\!\in\!\mathcal{S}$ is the full state, $a_t\!\in\!\mathcal{A}$ the action, and $o_t\!\in\!\mathcal{O}$ a partial observation. The transition kernel is $P(s_{t+1}\!\mid s_t,a_t)$, the observation model (sensor and preprocessing pipeline) is $\Omega(o_t\!\mid s_t)$, $r:\mathcal{S}\times\mathcal{A}\!\to\!\mathbb{R}$ is the reward, $\gamma\!\in\![0,1)$ the discount factor, and $\rho_0$ the initial‑state distribution. In our setting, $o_t$ comprises proprioception and a robot‑centric heightmap derived online from LiDAR, while $s_t$ additionally includes privileged quantities used only during training.
A stochastic policy $\pi_\theta(a_t\!\mid o_t)$ maximizes the discounted return
\begin{align}
J(\pi_\theta)=\mathbb{E}_{\substack{s_0\sim\rho_0\\
o_t\sim\Omega(\cdot\mid s_t),\,a_t\sim\pi_\theta(\cdot\mid o_t)\\
s_{t+1}\sim P(\cdot\mid s_t,a_t)}}\Bigg[\sum_{t=0}^{\infty}\gamma^t\,r(s_t,a_t)\Bigg].
\label{eq:objective}
\end{align}
\subsection{Robot-Centric Heightmap Representation}
Perception in PGTT relies on a compact representation of the terrain in the form of a \emph{robot-centric heightmap}. Unlike global elevation maps that require pose estimation and multi-sensor calibration, the heightmap is anchored to the robot's body frame and updated in real time from onboard LiDAR\footnote{We detail onboard LiDAR heightmap extraction in Sec.~\ref{sec:real}.} or using ground truth information in simulation.

\begin{figure}[h]
    % \vspace{-1em}
    \centering
    \includegraphics[width=0.8\linewidth]{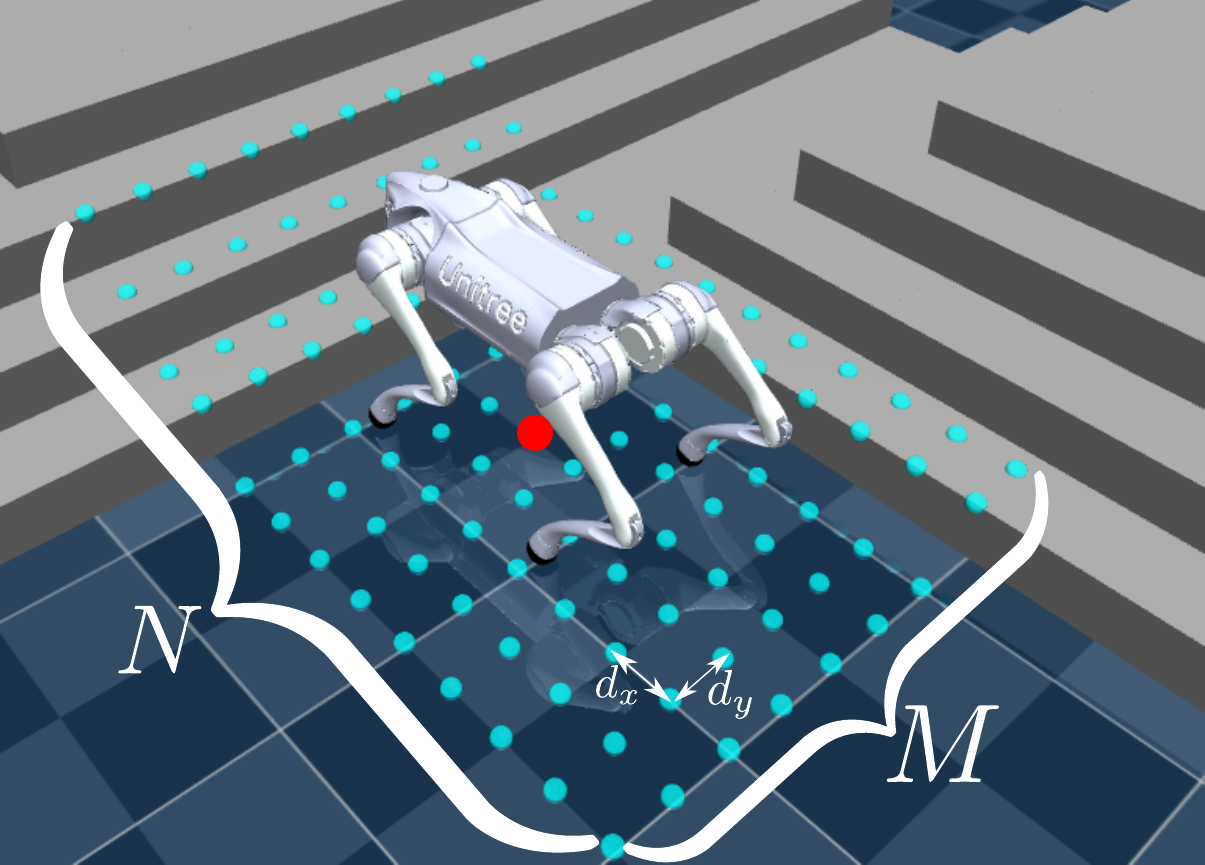}
    \caption{PGTT's robot-centric heightmap.}
    \label{fig:heightmap}
    \vspace{-1.5em}
\end{figure}

To define the heightmap, we create an $N\times M$ grid of equally spaced points around the robot (Fig.~\ref{fig:heightmap}), and pass it as input to the policy. Because the representation is local and robot-centric, it captures the geometry relevant to both front and hind legs without requiring global localization.

\subsection{Encoding Rhythmic Structure}
Following prior work that leverages rhythmic structure~\cite{Miki2022,Lee2020Learning,Shao_2022}, we define for each leg $i$ a periodic phase variable $\phi_{i,t}\in[0,2\pi)$, interpreted as \emph{contact} for $\phi_{i,t}\!\in\![0,\pi)$ and \emph{swing} for $\phi_{i,t}\!\in\![\pi,2\pi)$. The phase advances with a base frequency $f$ and over time $t$ as follows:
\begin{align}
\phi_{i,t} &= \big(\phi_{i,0} + 2\pi f t\big)\bmod 2\pi,
\label{eq:phase_update}
\end{align}
where $\phi_{i,0}$ sets the inter‑leg offsets (gait).
In PGTT, $\{\phi_{i,t}\}$ provides \emph{structure} but does not constrain the action space: it is used \emph{only} to shape the reward (see Sec.~\ref{sec:reward}) rather than to prescribe joint targets via oscillators or IK.
%
% \subsection{Action Space}
\subsection{Asymmetric Actor-Critic Learning}
Teacher–student distillation is a common recipe for locomotion with partial observations~\cite{Miki2022,Lee2020Learning}: a teacher is trained with full‑state (privileged) inputs and a student subsequently imitates it from observations. However, because the student is constrained to match the teacher’s demonstrations, the resulting policy can inherit suboptimality and distribution mismatch from the teacher’s occupancy measure. We therefore adopt \emph{asymmetric actor–critic}~\cite{pinto2017asymmetricactorcriticimagebased}, where the actor $\pi_\theta(a_t\!\mid o_t)$ conditions only on observations while the critic is trained with privileged state $s_t$; this retains the benefits of privileged information for value estimation without constraining the learned behavior by imitation~\cite{nahrendra2023dreamwaqlearningrobustquadrupedal}.

\textbf{Action Space:} The action space is a $12 \times 1 $ vector,
$a_t$, corresponding to the desired joint angle of the robot. To
facilitate learning, we train the policy to infer the desired
joint angle around the robot's stand still pose. Hence, the robot's desired joint angles are computed as 
\begin{align}
    q_{des}=q_{\text{stand}}+k a_t,
\end{align}
where $k$ is a constant \textit{action scale} parameter.

% \subsection{Observation Space}
\textbf{Observation Space:}
% \paragraph{Policy Network}
The observation space $o_t$, which is passed to the policy network $\pi_\theta(a_t|o_t)$, consists of mainly proprioceptive and exteroceptive measurements.
To encode the leg phase, we use
$\cos(\phi), \sin(\phi)$ instead of $\phi=[\phi_0,\phi_1,\phi_2,\phi_3]$, which is a smooth and unique representation for the angle \cite{Miki2022}. 
\begin{align}
    o_t=[\omega_t \ g_t \ q_t \ \dot{q_t} \ \cos(\phi) \ \sin(\phi) \ h_t \ f \ a_{t-1} \ v_{cmd}]^T,
\end{align}
where $\omega_t,g_t,q_t,\dot{q_t},\cos(\phi),\sin(\phi),h_t,f,a_{t-1} \  \text{and} \ v_{cmd}$ are the body angular velocity,
the gravity vector expressed in the local frame, the joint angles, the joint velocities, the phase representation, the flattened height-scans of the terrain, the base frequency, the last action and the command.

\textbf{Value Network:}
The value network is trained to output
an estimation of the true state value, $V(s_t)$. Unlike the policy the state $s_t$ contains privileged information $s_t=[o_t \ v_t \ ]^T$,
where $v_t$ is the local frame linear velocity.
Linear velocity is crucial because it correlates strongly with the main objective-track commanded velocity and thus with the value function output.

\subsection{Phase-Guided Reward Function}
\label{sec:reward}

Reward design is central to legged locomotion with reinforcement learning. Most existing approaches combine a forward-velocity tracking term with a set of penalties (slip, foot clearance) to promote stable gaits. While effective, these reward structures often require extensive manual tuning and are usually combined with oscillators or IK-based controllers.

PGTT pursues a different route: we aim to generate phase-guided swing trajectories \emph{without} inverse kinematics. The phase prior influences learning only through the reward, which reduces the number of hand-tuned terms and avoids constraining the policy. The core idea is to use \emph{cubic Hermite splines} to define smooth foot trajectories conditioned on a per-leg phase variable and local terrain information.

\begin{figure}[h]
    \vspace{-1em}
    \centering
    \includegraphics[width=.6\linewidth]{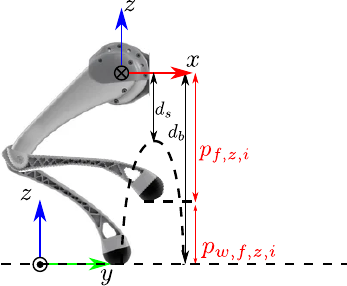}
    \caption{Distances relative to the hip-joint frame and world frame.
             The \textit{black} leg is the nominal stance, the 
             \textit{dashed} line a possible swing trajectory, and the 
             \textit{red} leg a random leg configuration.}
    \label{img:dist}
    \vspace{-1em}
\end{figure}

We denote by $p_{f,z,i}$ the $z$-axis (height) position of foot $i$ in the hip-joint frame, and by $p_{w,f,z,i}$ the corresponding position in the world frame. Let $d_b$ be the nominal foot height in stance (default configuration) and $d_s$ the nominal swing apex (see Fig.~\ref{img:dist}). To adapt the trajectory to terrain, we compute local statistics around each leg: $H_{\max,i}(h_t)$ and
$H_{\min,i}(h_t)$ are the maximum and minimum terrain heights in the world frame, and $\delta H_i(h_t) = H_{\max,i}(h_t) - H_{\min,i}(h_t)
$ is added to the swing trajectory to guarantee obstacle clearance.

Formally, a cubic Hermite spline is a map
$P(\,\cdot\,;\, p_0, p_1, m_0, m_1, T)$ parameterized by start and end
positions $p_0, p_1$, start and end tangents $m_0, m_1$, and duration $T$.
The independent variable and all parameters are expressed in phase
coordinates (radians of $\phi_{i,t}\in[0,2\pi)$) rather than in seconds.
For a local phase argument $\tau \in [0,T]$, the trajectory is
\begin{equation}
% \begin{aligned}
%     P(\tau) &= c_0 + c_1 \tau + c_2 \tau^2 + c_3 \tau^3, \\
%     c_0 &= p_0, \quad c_1 = m_0, \\
%     c_2 &= \tfrac{3}{T^2}(p_1-p_0) - \tfrac{2}{T}m_0 - \tfrac{1}{T}m_1, \\
%     c_3 &= -\tfrac{2}{T^3}(p_1-p_0) + \tfrac{1}{T^2}(m_0+m_1),
% \end{aligned}
\begin{aligned}
    P(\tau) &= c_0 + c_1 \tau + c_2 \tau^2 + c_3 \tau^3, \quad c_0 = p_0, \quad c_1 = m_0,\\
    c_2 &= \tfrac{3}{T^2}(p_1-p_0) - \tfrac{2}{T}m_0 - \tfrac{1}{T}m_1, \\
    c_3 &= -\tfrac{2}{T^3}(p_1-p_0) + \tfrac{1}{T^2}(m_0+m_1),
\end{aligned}
\end{equation}
so that $P(0)=p_0$, $P(T)=p_1$, $P'(0)=m_0$, and $P'(T)=m_1$.
We divide each leg trajectory into three phases:
% (parameterized by $\phi_{i,t}$):
\begin{itemize}
    \item \textbf{Stance:} the foot remains at $d_b$ for
    $0 \leq \phi_{i,t} < T_{\text{stance}}$, where
    $T_{\text{stance}} = 2\pi\, p_{\text{stance}}$ and $p_{\text{stance}}$
    is the stance ratio.
    \item \textbf{Swing up:} the swing-up spline, defined as
    \begin{equation}
    P_{su,i}(\tau, h_t) := P\big(\tau;\, d_b,\, d_s + \delta H_i(h_t),\, 0,\, 0,\, T_{\text{swing}}\big),
    \end{equation}
    with duration $T_{\text{swing}} = \pi(1-p_{\text{stance}})$.
    \item \textbf{Swing down:} the swing-down spline, defined as
    \begin{equation}
    P_{sd,i}(\tau, h_t) := P\big(\tau;\, d_s + \delta H_i(h_t),\, d_b,\, 0,\, 0,\, T_{\text{swing}}\big),
    \end{equation}
    beginning at $\phi_{i,t}=T_{\text{peak}}=\pi(1+p_{\text{stance}})$.
\end{itemize}
The desired $z$-position of foot $i$ at phase $\phi_{i,t}$ is then
\begin{equation}
\footnotesize
p^{\text{des}}_{f,z,i}(\phi_{i,t}, h_t)=
\begin{cases}
d_b, & 0 \leq \phi_{i,t} < T_{\text{stance}}, \\
P_{su,i}(\phi_{i,t} - T_{\text{stance}}, h_t), & T_{\text{stance}} \leq \phi_{i,t} < T_{\text{peak}}, \\
P_{sd,i}(\phi_{i,t} - T_{\text{peak}}, h_t), & T_{\text{peak}} \leq \phi_{i,t} < 2\pi.
\end{cases}
\label{eq:leg_traj}
\end{equation}
This compact definition provides structured yet terrain-adaptive trajectories that encourage swing clearance while leaving the policy free to discover joint-space behaviors. Reward terms penalize deviation from $p^{\text{des}}_{f,z,i}$ and swing-phase ground contacts, thereby shaping gait without explicit action-space constraints.
In practice, the overall reward is the sum of a small number of terms.  
Apart from the task-specific rewards (e.g. linear velocity tracking in our case), the central \emph{positive} term encourages each foot to follow its terrain-adaptive, phase-guided trajectory:
\begin{equation}
r_{\text{phase}} = \sum_{i \in \text{feet}} 
\exp\!\left(-\frac{\big(p^{\text{des}}_{f,z,i}(\phi_{i,t},h) - p_{f,z,i}\big)^2}{\sigma_f}\right).
\end{equation}

To discourage premature contacts during swing, we include a \emph{negative} penalty:
% \begin{equation}
$r_{\text{contact}} = \sum_{i \in \text{feet}} \mathds{1}_{\pi \leq \phi_{i,t} < 2\pi}\,c_i$,
% \end{equation}
where $c_i=1$ if foot $i$ is in ground contact and $0$ otherwise.  
This term penalizes collisions when the phase variable indicates that the leg should be swinging.
The final reward for the RL objective
% is a sum of many sub-rewards, and
can be found along with their weights in Table~\ref{table:rewards}.
\subsection{Terrain Generation}
To enable robust legged locomotion, we train policies in stair-like environments that capture the structure of indoor stairs while generalizing to boxy obstacles and irregular terrain. We generate these environments using the \textit{Wave Function Collapse} (WFC) algorithm, a constraint-satisfaction procedure originally introduced in~\cite{Gumin_Wave_Function_Collapse_2016}. WFC treats the environment as a grid where each cell can take on values (tiles) consistent with local adjacency rules. By iteratively ``observing'' one cell and propagating its constraints to neighbors, the algorithm produces diverse yet feasible layouts.

\begin{figure}[h]
    \centering
    \includegraphics[width=.9\linewidth]{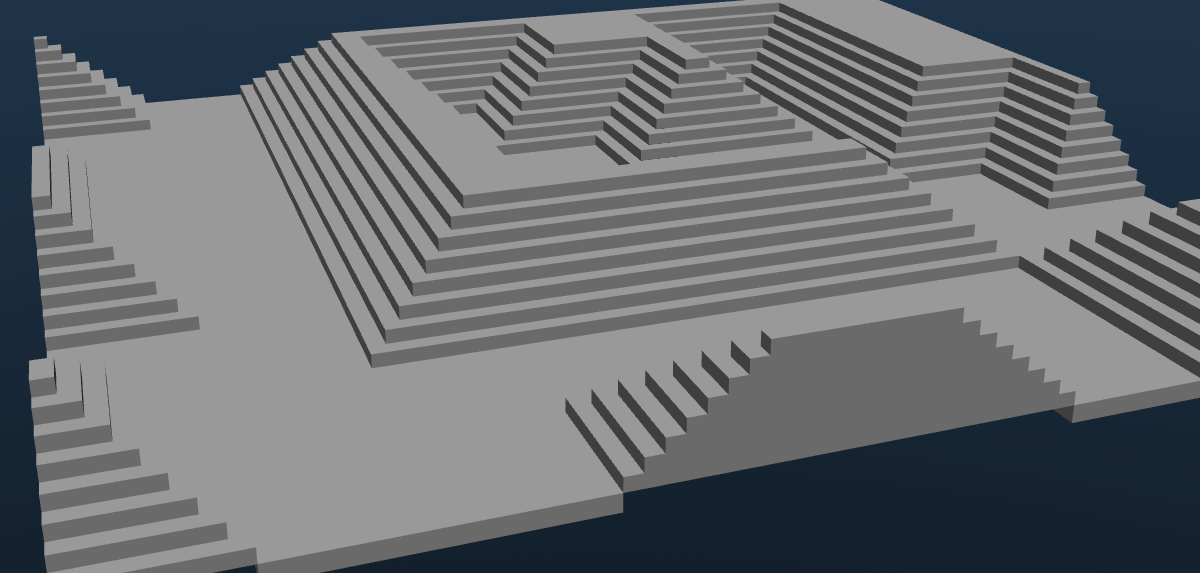}
    % \caption{Example of a generated environment with $w=0.1$, $h=0.08$, $n=8$.}
    \caption{Example of a procedurally generated stair-like environment.}
    \label{fig:stair_env}
    \vspace{-2em}
\end{figure}

Rather than primitive geometric tiles, we define higher-level architectural units (straight stair segments, corners, and flat floor tiles) as the building blocks. This representation encourages structural realism while still allowing variability. Each environment is represented as a two-dimensional grid of size $(2N{+}1)\times(2N{+}1)$, with the robot always spawning on the central tile at position $(N,N)$. Although the tiles are three-dimensional structures in simulation, the grid representation simplifies WFC’s operation while maintaining spatial consistency. Fig.~\ref{fig:stair_env} shows an example.
\subsection{Curriculum Learning}\label{sec:curr}
Direct training on highly irregular or steep terrains can impede convergence and result in brittle behaviors. To mitigate this, we adopt a staged curriculum in which terrain difficulty increases progressively. The curriculum comprises four levels of stair-like environments:
\begin{itemize}
    \item\textbf{Level 1:} nearly flat steps with minimal elevation changes,
    \item\textbf{Level 2:} moderate step heights,
    \item\textbf{Level 3:} tall obstacles requiring clear swing trajectories,
    \item\textbf{Level 4:} the highest steps the robot can safely traverse.%, designed to fully test gait robustness.
\end{itemize}
Training begins on the easiest terrain and advances only once the agent demonstrates reliable performance at each level. By mastering balance and stepping on low stairs before tackling larger obstacles, the policy gradually acquires the necessary stability and clearance behaviors. This staged progression improves sample efficiency and results in policies that transfer more reliably to challenging real-world scenarios.

\subsection{Sim-to-Real Transfer}
To improve robustness and bridge the sim-to-real gap, we apply extensive domain randomization during training:%. More specifically:
\begin{itemize}
  \item \textbf{Sensor noise:} Gaussian noise is injected into all components of the observation vector $o_t$ to mimic measurement uncertainty and encourage robustness to perception artifacts.
  \item \textbf{Robot properties:} We randomize key physical parameters such as link and torso masses, default joint positions, motor gains ($k_p, k_d$), and actuator friction, mitigating sensitivity to modeling errors.
  \item \textbf{Environment properties:} We vary terrain and stair friction coefficients to account for diverse contact conditions in the real world.
\end{itemize}
This randomization forces the policy to generalize across a wide range of conditions, improving stability and survivability in hardware deployment.
\subsection{Training Details}
We train legged locomotion policies in the MuJoCo physics simulator, using its GPU-accelerated JAX~\cite{jax} branch (MJX) to achieve high-throughput simulation~\cite{playground}.
Compared to the widely used Isaac Gym~\cite{makoviychuk2021isaacgymhighperformance}, MuJoCo/MJX requires significantly less computational power, making large-scale training of perceptive locomotion policies more accessible.
Our pipeline combines procedurally generated stair-like terrains, GPU-accelerated heightmap extraction, curriculum learning, and domain randomization. MuJoCo provides accurate contact dynamics, while MJX allows the full physics computation to run on the GPU. On top of MJX, we leverage \textsc{Brax}~\cite{brax},
which provides efficient hardware acceleration and clean integration with modern reinforcement learning algorithms. The policy is optimized using Proximal Policy Optimization (PPO)~\cite{schulman2017proximal}, a widely adopted on-policy algorithm.

\section{Experimental Setups and Results}\label{sec:exps}
In this section we will present the results of the proposed policy in simulation and the real world and compare them with strong baselines using several metrics.
All policies were trained on a workstation equipped with an Intel Core i9-14900K CPU and a single NVIDIA GeForce RTX 5070 GPU. Training used a physics-integration time step of \( dt = \SI{0.005}{s}\).  During deployment, in both Sim2Sim and Sim2Real transfers, control commands are issued at 50 Hz (i.e., every 0.02 s).
\begin{table}
\footnotesize
\centering
\scriptsize
\setlength{\tabcolsep}{3pt} % reduce horizontal spacing
\captionsetup{font=footnotesize, singlelinecheck=false}
\caption{Reward functions. First section contains common tasks, second section the rewards for \textit{PGTT}, third section the \textit{MassLoco} method rewards and fourth section the \textit{Wild} method rewards. The indicator function $\mathds{1}_c$ obtains the value 1 if c is true and is 0 otherwise.}
\vspace{-1em}
\label{table:rewards}
\begin{center}
\begin{tabular}{lll}
\hline \hline
Reward           & Equation ($r_i$) & Weight ($w_i$) \\ \hline 
Lin. velocity tracking & $\exp\left(-\frac{(\mathbf{v}^{\mathrm{cmd}}_{xy}-\mathbf{v}_{xy})^2}{\sigma_v}\right)$ & $1.0$ \\  
Ang. velocity tracking & $\exp\left(-\frac{(\omega^{\mathrm{cmd}}_{z}-\omega_{z})^2}{\sigma_v}\right)$ & $0.5$ \\
Linear velocity ($z$) & $v^2_z$ & $-2.0$ \\
Angular velocity ($xy$) & ${\omega}^2_{xy}$ & $-0.05$ \\
Orientation & $|\boldsymbol{g}|^2$ & $-0.2$ \\
Termination & $\mathds{1}_{termination\ early}$ & $-1.0$\\
% Joint accelerations & ${\ddot{\theta}}^2$ & $-2.5\!\times\!10^{-7}$ \\
Joint power & $|{\tau}||{\dot{\theta}}|$ & $-2\!\times\!10^{-5}$ \\
Action rate & $(a_t - a_{t-1})^2$ & $-0.01$ \\
Joint limits & $ \mathds{1}_{q_i>q_{max} || q_i < q_{min}}$ & $-1.0$\\
Default pose & $ \sum_{i \in \text{foot}}(q-q_{def})^2\cdot w_i$ & $-0.5$\\
Joint torques & $|\tau|^2$ & $-1\times10^{-5}$ \\
\hline
\textbf{PGTT}\\
Foot phase & $\sum_{i \in \text{foot}} \exp\left(-\frac{(p^{\text{des}}_{f,z,i}(\phi_{i,t},h)-p_{f,z,i})^2}{\sigma_f}\right)$ & $1.0$ \\
Foot contact & $\sum_{i \in \text{foot}} \mathds{1}_{\pi \leq \phi_{i,t} < 2\pi} c_i$ & $-0.25$\\
\hline
\textbf{MassLoco} \\
Foot clearance & $\sum_{i \in \text{foot}}(p^{\text{des}}_{w,f,z,i}-p_{w,f,z,i})^2\cdot \|v_{f,xy,i}\|_2 $& $-0.5$ \\
Foot slip & $\sum_{i \in \text{foot}} ( \|v_{f,xy,i}\|_2 \cdot c_i) $ & $0.1$\\
Feet air time & $\mathds{1}_{\|v_{cmd}\|_2>0.01} \sum_{i \in \text{foot}}(t_{i,air}-0.5) $ & $1.0$\\
Stand still & $\mathds{1}_{\|v_{cmd}\|_2<0.01}(q-q_{def}) $ & $0.5$ \\
\hline
\textbf{Wild} \\
Foot clearance & $\sum_{i \in \text{foot}}\mathds{1}_{\pi \leq \phi_{i,t} < 2\pi} \mathds{1}_{p_{w,f,z,i} \geq  H_{max,i}} $& $0.1$ \\
Foot slip & $\sum_{i \in \text{foot}} ( \|v_{f,xy,i}\|_2 \cdot c_i) $ & $0.1$\\
\hline \hline
\end{tabular}
\end{center}
% \begin{tablenotes}\footnotesize
% \item[*] 
% \end{tablenotes}
% \end{threeparttable}
\vspace{-3em}
\end{table}
\subsection{Baselines}
We select baseline methods that are both relevant to our problem and representative of existing approaches to enable meaningful evaluation.
To evaluate whether locomotion without fixed gait scheduling can yield more efficient behaviors, we include \textit{MassLoco}~\cite{rudin2022learningwalkminutesusing}, including rewards inspired by Margolis et al.~\cite{margolis2022walkwaystuningrobot} to encourage more natural walking patterns. On the other hand, when considering a state-of-the-art method that leverages gait priors, we compare against \textit{Wild}~\cite{Miki2022}. We did not include Visual CPG-RL~\cite{bellegarda2024visualcpgrllearningcentral}, since, although its framework is similar to Wild, it is not trained or evaluated on stairs or obstacle traversal, and is therefore considered less relevant for our study.

Table~\ref{table:rewards} summarizes the rewards used by our method and the baselines. To isolate the effect of the reward formulation, all three methods share an identical experimental setup and differ \emph{only} in their reward terms: they use the same observation space $o_t$ and privileged state $s_t$, the same actor and critic architectures and capacity and the same terrain generated environments. We include the ordinary rewards for velocity tracking (typically $\sigma_v=0.25$) and some regularization rewards to maintain balance, avoid early termination and excessively large stress of the joints; these shared rewards use common weights across all methods. The method-specific reward weights are taken from each baseline's original work
and adjusted only as needed for stable training under our setup, with comparable tuning effort across all
methods. We also include rewards to avoid joint limits, where $q_{max},q_{min}$ are soft limits for the joints of the robot, and a reward to penalize deviations from the default joint position with different weights for each joint.

For PGTT two additional rewards are necessary: \textit{foot phase} is responsible for tracking the desired leg trajectories (with a small $\sigma_f=0.05$), and \textit{foot contact} penalizes contacts at swing phase.
In contrast, the MassLoco method requires four additional reward terms. Foot clearance promotes high clearance strides,  since it penalizes deviation from the desired swing height when the feet move in the xy plane (non-zero velocity), foot slip penalizes contacts when feet move in the xy plane, feet air time promotes long strides when a command is given and stand still promotes maintaining the default configuration when no command is given.
The Wild method uses the same foot slip reward as MassLoco, but its foot clearance reward specifically encourages swings that rise higher than the surrounding obstacles.
\begin{figure*}[t]
    \centering
    \begin{subfigure}[b]{0.32\textwidth}
        \includegraphics[width=\textwidth]{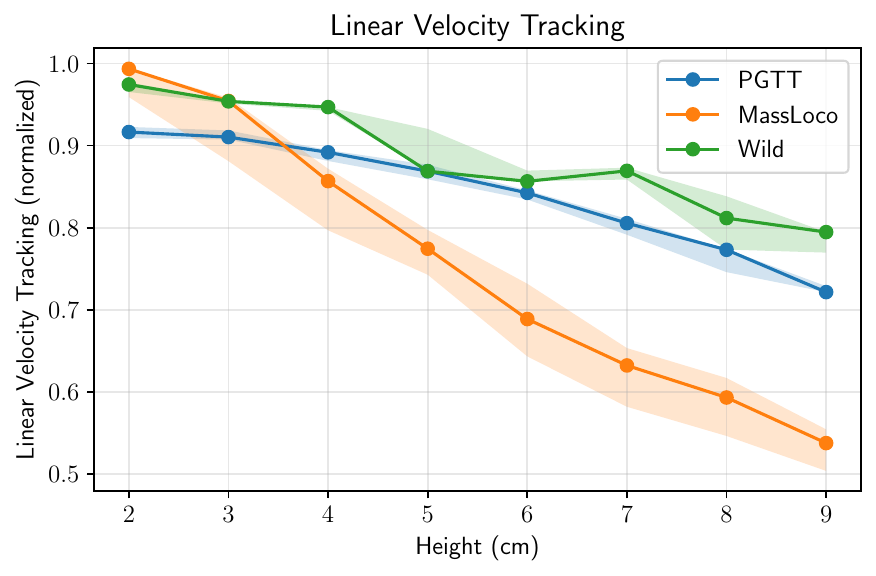}
        \caption{Linear velocity}
        \label{fig:linear_velocity}
    \end{subfigure}
    \hfill
    \begin{subfigure}[b]{0.32\textwidth}
        \includegraphics[width=\textwidth]{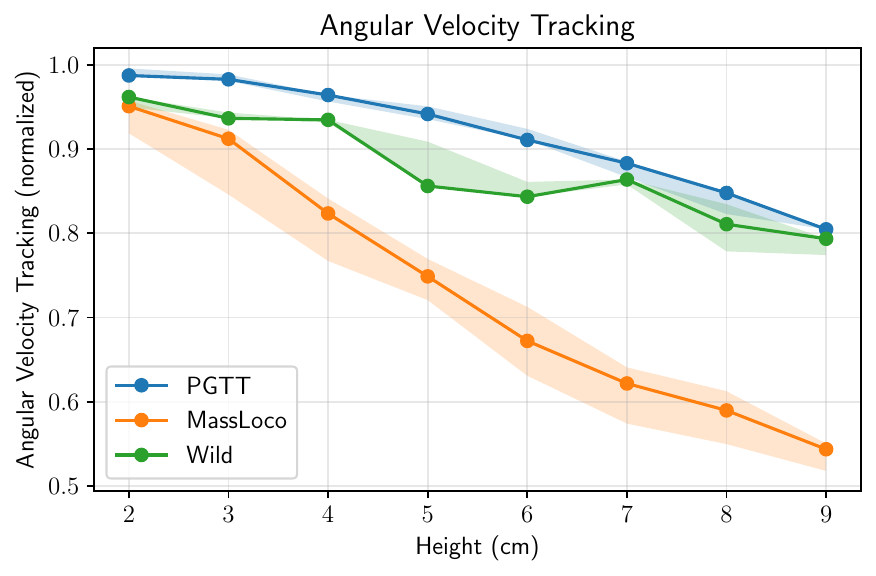}
        \caption{Angular velocity}
        \label{fig:angular_velocity}
    \end{subfigure}
    \hfill
    \begin{subfigure}[b]{0.32\textwidth}
        \includegraphics[width=\textwidth]{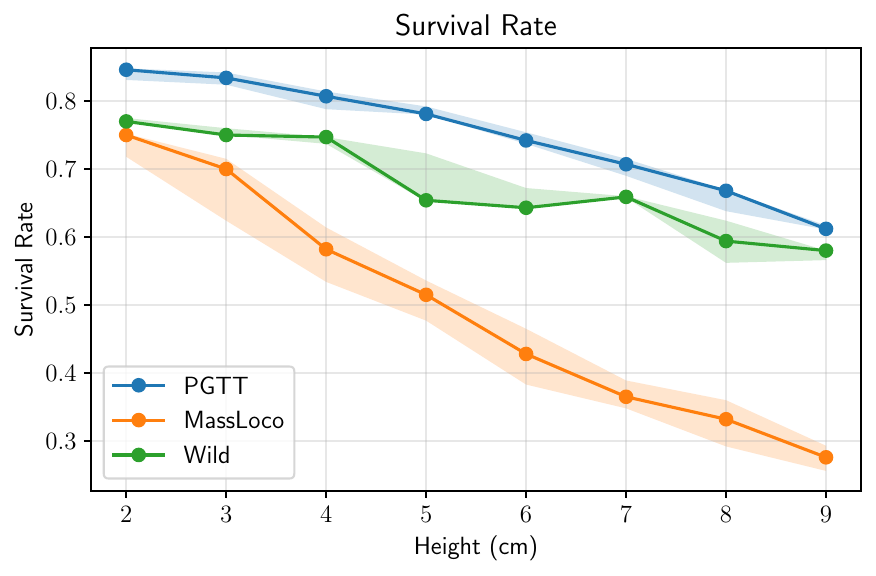}
        \caption{Success rate}
        \label{fig:success_rate}
    \end{subfigure}

    \caption{Comparison of PGTT with baseline methods MassLoco and Wild across three metrics: linear velocity tracking, angular velocity tracking and success rate. \textit{Solid lines} show the median over 5 different training seeds and the shaded regions are the regions between the 25-th and 75-th percentiles.}
    \label{fig:comparison_all_metrics}
    \vspace{-1em}
\end{figure*}
\subsection{Implementation Details}
At the start of each episode we draw a constant command
\(v_{cmd} = [v_x\; v_y\; \omega_z]^T \) uniformly from the
intervals \(u_{min}=(-1,-1,-1)\) and
\(u_{max}=(1,1, 1)\).  To expose the policy to discontinuous
command changes, we resample \(v_{cmd}\) once per episode at a
random time-step. Additionally, we sample frequencies $f\sim U[1,3]$ to create diverse gaits in terms of timings.

The actor and critic networks are modeled as multilayer perceptrons (MLPs) with hidden layer sizes of 512, 256, and 128. Our choice of using an MLP with elevation maps is justified, as prior work has shown that memory mechanisms are not required for this modality, with both MLP and LSTM architectures achieving comparable reconstruction performance~\cite{rudin2025parkourwildlearninggeneral}.
Episodes are terminated early if the robot turns upside down to further accelerating training.
\subsection{Curriculum Learning}
We employ Curriculum Learning to structure the training process as detailed in Sec.~\ref{sec:curr}. At all levels we use a grid size of 5 (i.e., $2N+1=5$), step width is sampled as $ w \sim U[0.3,0.45]$, number of steps are sample as $ n\sim U[2,4]$. We sample the height (in cm) of the terrain differently per level: between $[1,3]$ in level 1, $[1,7]$ in level 2, $[1,10]$ in level 3, and $[1,13]$ in level 4.
% ; the only difference is the height, where in level 1 we sample from 1cm to 3cm , in level 2 from 1cm to 7cm, for level 3 we sample from 1cm to 10cm and fore level 4 from 1cm to 13cm.
These choices allow for efficient learning and overcoming progressively higher obstacles, while retaining the ability to traverse smaller ones.

Determining when the agent has successfully completed a level, however, is non-trivial. 
To address this, we utilize a velocity tracking metric to assess level completion. 
Specifically, during each evaluation step, using 
$n_{eval}$ agents, we compute the cumulative velocity reward as follows:
% \begin{equation}
\begin{align}
  m_v=& \frac{1}{n_{eval} T} \frac{1}{w_v} \sum_{e=1}^{n_{eval}} \sum_{t=0}^{T-1} w_v e^{-\frac{(v_t-v_{cmd})^2}{\sigma}} \label{eq:vel}\nonumber\\
  m_\omega=& \frac{1}{n_{eval} T} \frac{1}{w_\omega} \sum_{e=1}^{n_{eval}} \sum_{t=0}^{T-1} w_\omega e^{-\frac{(\omega_t-\omega_{cmd})^2}{\sigma}} 
\end{align}
% \end{equation}

Therefore, we define success of a level if $m_v,m_\omega \geq p$, where $p \in [0,1)$. We set $p=0.65$. Additionally, we will declare convergence only when the episode reward $R_t$ has stabilized, that is, $\frac{|R_t-R_{t-1}|}{R_{t-1}}<\epsilon$.
% \begin{equation}
%   \frac{|R_t-R_{t-1}|}{R_{t-1}}<\epsilon  
% \end{equation}
%
\subsection{Metrics}
We compare our method PGTT with the two aforementioned baselines
% MassLoco and Wild
in terms of linear and angular velocity tracking, and success rate. The first two are defined as in Table~\ref{table:rewards}.
Success rate (SR) is defined
% as follows~\textit{Gangapurwala et al.}
as (following~\cite{gangapurwala2023learninglowfrequencymotioncontrol}): $\text{SR} = 1 - N_e/N_T$,
% :
% \begin{equation}
%     \text{SR} = 1 - \frac{N_e}{N_T},
% \end{equation}
with $N_e$ referring to the number of rollouts that terminated early due to a prohibited behavior and $N_T$ being the total number of rollouts. Using $N_T=1000$, we randomize the base linear and angular velocity command  with $0.7 \cdot v^\max$ from the one used during training.

\subsection{Simulation Results}
We compared the three methods in generated stair environments with obstacle heights ranging from 2cm to 9cm. While the robot can traverse obstacles up to 12cm, the most stable behaviors were observed at heights between 7cm and 9cm. We would like to compare the methods when perturbations are applied to the robot torso towards any direction, to validate the robustness in realistic scenarios. We apply perturbations of uniformly sampled magnitude between 7.5 to 30 N and we also sample durations and wait times between consecutive perturbations. We replicate the whole training pipeline over 5 different seeds. All metrics excluding the success rate are normalized with respect the the maximum value. The results reveal several clear trends (Fig.~\ref{fig:comparison_all_metrics}): PGTT and Wild exhibit very similar commanded velocity tracking, whereas MassLoco lags behind. PGTT achieves the highest success rate, \textbf{outperforming the second-best method, Wild, by 7.5\% on average}. 

Similar behavior is observed when evaluating the three methods in environments with discrete obstacles, with PGTT achieving the highest success rate; \textbf{9\% higher than the second-best method}. Additionally, both angular and linear velocity tracking are very similar across all methods (Table~\ref{table:ablation1}).

\begin{table}[t]
\centering
\caption{
Evaluation metrics when the robot traverses discrete obstacles of varying height from 2cm to 9cm.
Success rate, normalized body linear velocity error $\bar{v}$, and normalized body angular velocity error $\bar{\omega}$ for 1000 quadrupeds. Results are (\textbf{median}, 25th percentile, 75th percentile) over 5 training seeds.
}

\renewcommand{\arraystretch}{2.5} % row height

\resizebox{\columnwidth}{!}{
\begin{tabular}{ | c || c | c | c |}
\hline
Method & \thead{Success\\ Rate} & $\bar{v}$ & $\bar{\omega}$  \\
\hline
PGTT      
 & \cellcolor{cellgreen}$(\mathbf{0.848},0.842,0.855)$ 
 & \cellcolor{cellred}$(\mathbf{0.965},0.958,0.972)$ 
 & \cellcolor{cellgreen}$(\mathbf{0.991},0.986,0.994)$ \\
\hline
MassLoco 
 & \cellcolor{cellred}$(\mathbf{0.702},0.659,0.711)$ 
 & \cellcolor{cellorange}$(\mathbf{0.983},0.939,0.986)$ 
 & \cellcolor{cellred}$(\mathbf{0.903},0.863,0.904)$ \\
\hline
Wild 
 & \cellcolor{cellorange}$(\mathbf{0.756},0.756,0.769)$ 
 & \cellcolor{cellgreen}$(\mathbf{0.998},0.998,1.000)$ 
 & \cellcolor{cellorange}$(\mathbf{0.935},0.935,0.941)$ \\
\hline
\end{tabular}
}
\label{table:ablation1}
\vspace{-1em}
\end{table}

\begin{figure}[h]
    \centering
    \includegraphics[width=0.9\linewidth]{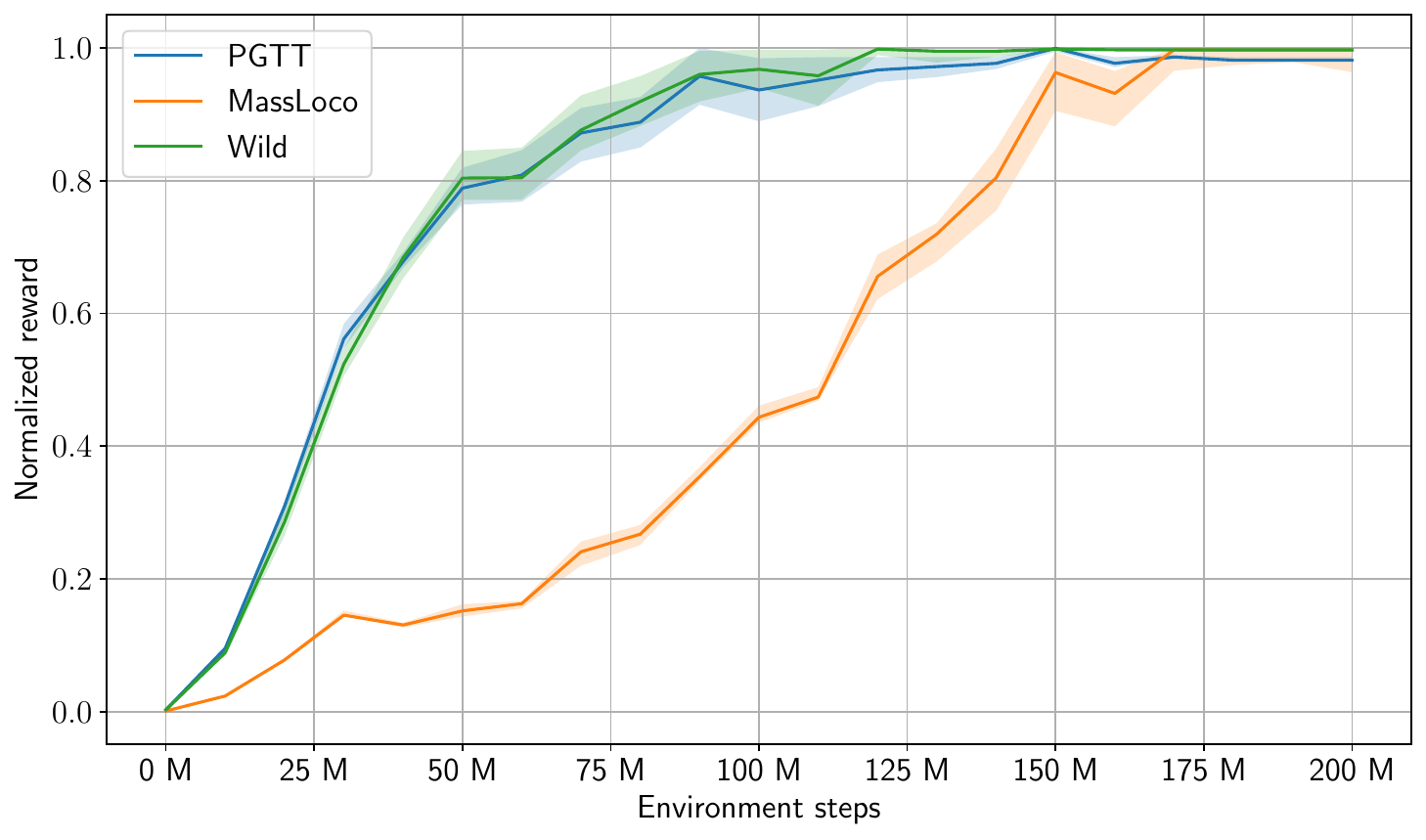}
    \caption{Training curves. We report only for Level 1 of our curriculum learning, since this is the level with the biggest differences. \textit{Solid lines} show the median over 5 different training seeds and the shaded regions are the regions between the 25-th and 75-th percentiles.}
    \label{fig:training_time}
    \vspace{-1em}
\end{figure}

In terms of convergence of the policy learning, PGTT and Wild are mostly equivalent, whereas MassLoco is considerably less sample-efficient, needing about twice as many steps in the first curriculum level (Figure~\ref{fig:training_time}). In terms of wall-time, the mean completion time (averaged over five training seeds) across all four levels are 195, 198, and 239 minutes for PGTT, Wild, and MassLoco, respectively.
\subsection{Ablation Study}
To assess the contribution of PGTT's two additional reward terms, we compare the full method against two ablations: \textit{w/o foot phase}, which removes the foot phase tracking reward , and \textit{w/o foot contact}, which removes the foot contact penalty. All other rewards, architecture, and training setup are kept identical to PGTT. Results are evaluated on stairs with a step height of 7cm, chosen as a representative baseline difficulty.
\begin{table}[t]
\centering
\caption{
Ablation of PGTT's reward components on stairs (7cm step height).
Success rate, normalized body linear velocity error $\bar{v}$, and normalized body angular velocity error $\bar{\omega}$ for 1000 quadrupeds. Results are (\textbf{median}, 25th percentile, 75th percentile) over 5 training seeds.
}
\renewcommand{\arraystretch}{2.5}
\resizebox{\columnwidth}{!}{
\begin{tabular}{ | c || c | c | c |}
\hline
Method & \thead{Success\\ Rate} & $\bar{v}$ & $\bar{\omega}$  \\
\hline
PGTT (full)
 & $(\mathbf{0.834},0.831,0.838)$
 & $(\mathbf{0.987},0.979,0.996)$
 & $(\mathbf{0.989},0.982,0.997)$ \\
\hline
w/o foot phase
 & $(\mathbf{0.755},0.749,0.762)$
 & $(\mathbf{0.992},0.987,0.998)$
 & $(\mathbf{0.997},0.994,0.999)$ \\
\hline
w/o foot contact
 & $(\mathbf{0.826},0.820,0.832)$
 & $(\mathbf{0.995},0.994,0.999)$
 & $(\mathbf{0.996},0.993,0.998)$ \\
\hline
\end{tabular}
}
\label{table:ablation_components}
\vspace{-1.5em}
\end{table}

Table~\ref{table:ablation_components} shows that the foot phase reward is the main driver of PGTT's success rate on stairs: removing it drops success rate from \textbf{83.4\%} to \textbf{75.5\%}, while velocity tracking remains largely unaffected. The foot contact penalty, by contrast, contributes a smaller refinement on top of this, improving success rate from \textbf{82.6\%} to \textbf{83.4\%}; it was primarily added for extra stability in edge cases, e.g., when the swing foot brushes against a step edge before completing its trajectory. Finally, PGTT's slightly worse velocity tracking reflects the robot prioritizing balance over speed while on the stairs, not a real drop in performance.

\textbf{ANYmal C Experiments:} We also applied our method on the same task with the ANYmal C robot. Our preliminary experiments showcase that PGTT is able to generate walking behaviors without even changing the hyper-parameters (see Fig.~\ref{fig:simulation} and the supplementary video\footnote{Available also at \url{https://youtu.be/v3iIu6ceFDM}.}).
\subsection{Real-World Deployment}\label{sec:real}
\begin{figure}[tb]
    \centering
    \includegraphics[width=1.0  \linewidth]{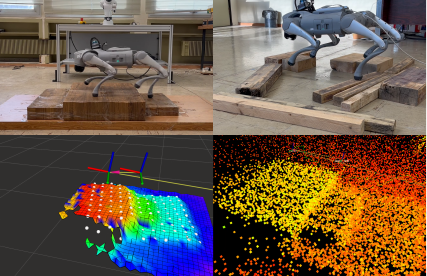}
    \caption{Real world experiments. The bottom left image shows the gridmap and the bottom right shows the odometry.}
    \label{fig:go2_real}
    \vspace{-2em}
\end{figure}
We evaluate the Sim2Real capabilities of our method on a Unitree Go2 quadruped. For perception, a L1 LiDAR is fused with IMU data using Point-LIO~\cite{He2023PointLIO}, a tightly coupled LiDAR–Inertial Odometry framework. The resulting odometry and transformed point cloud are used to construct a robot-centric elevation grid map~\cite{Fankhauser2018ProbabilisticTerrainMapping}, where each cell $(i,j)$ stores a mean height $\hat{h}_{ij}$ and variance $\sigma_{ij}^2$ to represent terrain uncertainty. Since raw LiDAR maps often contain holes
that can destabilize the policy, we apply a median-fill filter that in-paints only small gaps (below radius $r_{\text{hole}}$) surrounded by reliable data, while leaving larger unknown regions untouched.

To provide real-time input to the policy, we extract a robot-centric heightmap by sampling an $11\times 9$ grid within a $1.1\,m\times 0.9\,m$ area centered on the robot. Although accuracy is bounded by grid resolution, the domain randomization used during training makes the policy robust to such imperfections. The locomotion policy executes at $50\,Hz$, producing joint targets that are translated into torques through a lightweight PD controller ($k_p=60$, $k_d=3$) before being applied by the Go2's onboard low-level controller.

Our experiments showcase that policies trained with the PGTT effectively transfer to the real-world (Fig.~\ref{fig:go2_real} and supplementary video), and the robot is able to walk both on static stair and discrete obstacles environments, and withstand real-life perturbations.

\section{Conclusion}
In this work, we introduced \textbf{Phase-Guided Terrain Traversal (PGTT)}, a perception-aware locomotion framework that combines local heightmap perception, reinforcement learning, and terrain-adaptive gait priors for robust terrain traversal. PGTT improves traversal success by \textbf{7.5\%}. Preliminary results on ANYmal-C suggest cross-platform generalization without inverse kinematics, while reliable deployment on a Unitree Go2 demonstrates transfer to real hardware. PGTT also relies on the lightweight MuJoCo simulation stack, enabling perception-aware policies to be trained on affordable hardware, including a single consumer-grade GPU, and lowering the barrier to entry for this research.

At the same time, our approach has limitations. The L1 LiDAR used in our implementation outputs only 21{,}600 points/s, considerably fewer than higher-end sensors ($\geq$200{,}000 points/s). At higher locomotion speeds, this leads to sparser measurements and less frequent map updates, restricting our experiments to a maximum speed of $0.4\,m/s$. This issue, however, is tied to sensing hardware and could be readily addressed with improved sensors. Another limitation is that we have not evaluated the energy efficiency of PGTT in real-world deployments; understanding its effect on power consumption will be important for scaling to long-duration or field applications.

Looking forward, because PGTT's phase prior is expressed only through the reward rather than the action space, it is a natural candidate for training a single policy across multiple morphologies simultaneously, for instance by conditioning the reward on morphology-specific parameters while sharing the same action space, which we leave for future work.

Overall, PGTT provides an accessible and effective foundation for advancing agile, robust, and affordable legged locomotion in real-world environments, empowering researchers and laboratories without extensive computational resources to contribute to this field.

%%%%%%%%%%%%%%%%%%%%%%%%%%%%%%%%%%%%%%%%%%%%%%%%%%%%%%%%%%%%%%%%%%%%%%%%%%%%%%%%
% \vspace{-0.7em}
\bibliographystyle{ieeetr}
\bibliography{thesis}

\end{document}